\documentclass[numbers]{article}

\usepackage[final]{neurips_2024}

\usepackage[utf8]{inputenc} 
\usepackage[T1]{fontenc}    
\usepackage{hyperref}       
\usepackage{url}            
\usepackage{booktabs}       
\usepackage{amsfonts}       
\usepackage{nicefrac}       
\usepackage{microtype}      
\usepackage{xcolor}         
\usepackage{graphicx}
\usepackage{booktabs}
\usepackage{multirow}
\usepackage{colortbl}
\usepackage{array}
\usepackage{amsmath}

\usepackage{amssymb}

\title{DARD: \\A Multi-Agent Approach for \\Task-Oriented Dialog Systems}

\author{%
  Aman Gupta\thanks{Work performed during an internship at Amazon} \\
  Carnegie Mellon University\\
  \texttt{amangupt@cs.cmu.edu} \\
  \And
  Anirudh Ravichandran \\
  Amazon \\
  \texttt{aniravic@amazon.com} \\
  \And
  Ziji Zhang \\
  Amazon \\
  \texttt{czhangzi@amazon.com} \\
  \And
  Swair Shah \\
  Amazon \\
  \texttt{shahswai@amazon.com} \\
  \And
  Anurag Beniwal \\
  Amazon \\
  \texttt{beanurag@amazon.com} \\
  \AND
  Narayanan Sadagopan \\
  Amazon \\
  \texttt{sdgpn@amazon.com} \\
}

\begin{document}



\maketitle

\begin{abstract}
Task-oriented dialogue systems are essential for applications ranging from customer service to personal assistants and are widely used across various industries. However, developing effective multi-domain systems remains a significant challenge due to the complexity of handling diverse user intents, entity types, and domain-specific knowledge across several domains. In this work, we propose DARD (Domain Assigned Response Delegation), a multi-agent conversational system capable of successfully handling multi-domain dialogs. DARD leverages domain-specific agents, orchestrated by a central dialog manager agent. Our extensive experiments compare and utilize various agent modeling approaches, combining the strengths of smaller fine-tuned models (Flan-T5-large \& Mistral-7B) with their larger counterparts, Large Language Models (LLMs) (Claude Sonnet 3.0). We provide insights into the strengths and limitations of each approach, highlighting the benefits of our multi-agent framework in terms of flexibility and composability. We evaluate DARD using the well-established MultiWOZ benchmark, achieving state-of-the-art performance by improving the dialogue inform rate by 6.6\% and the success rate by 4.1\% over the best-performing existing approaches. Additionally, we discuss various annotator discrepancies and issues within the MultiWOZ dataset and its evaluation system. 
\end{abstract}

\section{Introduction}

In recent research, significant efforts have been made to build systems that involve planning and communication between various specialized agents to perform complex tasks \cite{yao2023reactsynergizingreasoningacting, liu2023agentbenchevaluatingllmsagents, liang2024encouragingdivergentthinkinglarge}. These agents are, in turn, backed by instruction-tuned open-source LLMs, external APIs, or other simpler tools. Various tasks such as logical reasoning \cite{du2023improving, tang2024medagentslargelanguagemodels}, societal simulations \cite{zhou2024sotopiainteractiveevaluationsocial, kaiya2023lyfeagentsgenerativeagents}, software development \cite{hong2023metagptmetaprogrammingmultiagent} have seen remarkable improvement in performance using these multi-agent framework methods. 

In this work, we explore the potential of agentic design in Task-Oriented Dialogue Systems (TODS). TODS are prevalent in real-world applications, such as customer service, e-commerce, and commercial voice assistants like Amazon Alexa and Google Assistant. Multiple research efforts have been made to curate high-quality labeled datasets to aid in developing systems that can handle end-to-end task-oriented dialogs \cite{budzianowski2018large, rastogi2020scalablemultidomainconversationalagents, henderson-etal-2014-second}. One of the most well-established and widely used TOD datasets amongst them is the MultiWOZ dataset \cite{budzianowski2018large}. The dataset contains 10k+ single and multi-domain conversations spanning 7 domains of attraction, hospital, hotel, restaurant, taxi, train, and police. After the original release, multiple corrected versions (2.1 - \cite{eric-etal-2020-multiwoz}, 2.2 - \cite{zang-etal-2020-multiwoz}, 2.3 - \cite{han2021multiwoz23multidomaintaskoriented}, 2.4 - \cite{ye-etal-2022-multiwoz}) of the MultiWOZ dataset have been published, each addressing specific issues. We primarily experiment with the MultiWOZ 2.2 \cite{zang-etal-2020-multiwoz} version, as it had the most established benchmark \cite{nekvinda-dusek-2021-shades} and is the latest recognized version as per the official repository\footnote{\url{https://github.com/budzianowski/multiwoz}}. The main tasks performed on the MultiWOZ dataset are Dialogue State Tracking (DST), which involves tracking predefined slots and their values in the context, and Response Generation, which involves predicting the system's response to the latest user message.

We propose a \textbf{D}omain \textbf{A}ssigned \textbf{R}esponse \textbf{D}elegation (\textit{DARD}), a framework that involves multiple specialized domain agents invoked by a central dialog manager agent, based on the dialog context. These domain-specific agents use conversational context and relevant entities from an external database to generate a response for the latest user message(see Figure \label{fig:overview}). We experiment with fine-tuned Flan-T5-Large \cite{chung2022scalinginstructionfinetunedlanguagemodels}, Mistral-7B \cite{jiang2023mistral7b}, and prompted Claude Sonnet 3.0 \footnote{\url{https://www.anthropic.com/news/claude-3-family}} models as our domain agents and also a prompted Claude Sonnet 3.0 as a dialog manager agent. We present further details of our experiments in the section \label{sec:methods}. We further systematically compare the performance of our approaches with other top-performing approaches on the benchmark in the section  \label{sec:results}. To rigorously validate our method and its performance, we conduct comprehensive error analyses and also present the challenges and limitations of the MultiWOZ dataset and its evaluation system in section \label{sec:discussions}. The following are the key contributions and insights we present through our work: 

\begin{figure*}[tp]
    \centering
    \includegraphics[width=\textwidth]{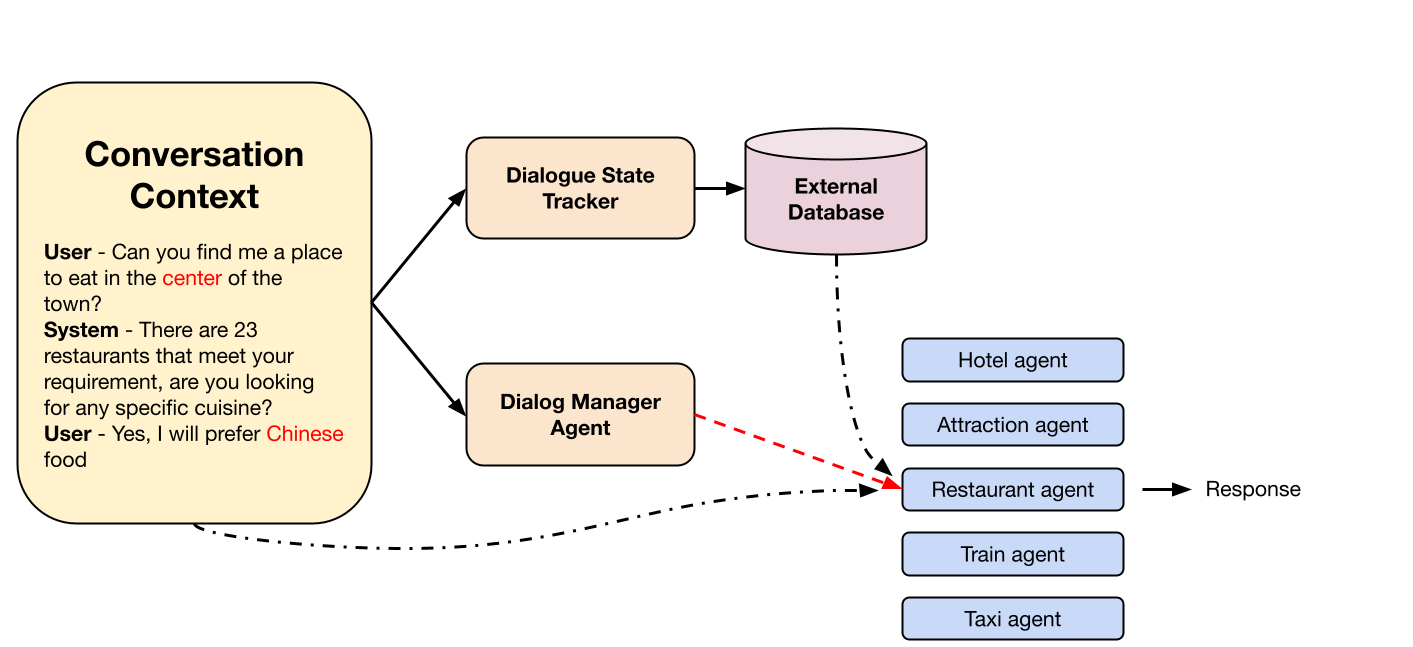}
    \caption{Overview of \textit{DARD} for end-to-end response generation on MultiWOZ. The current diagram shows a conversation assigned to the restaurant agent but in general, it can be assigned to any of the domain agents}
    \label{fig:overview}
\end{figure*}


\begin{itemize}
    \item We introduce DARD (Domain-Assigned Response Delegation), an ensemble of domain-specific agents that improve the state-of-the-art dialog inform rate by 6.6\% and success rate by 4.1\% on the MultiWOZ benchmark. 
    \item Our study presents a detailed comparison of performance between fine-tuned(Mistral-7B, Flan-T5-Large) vs Prompted (Claude Sonnet 3.0) models in the context of dialog agents and single-agent vs multi-agent approaches for task-oriented dialogs
    \item Our analysis reveals multiple annotator discrepancies present in the MultiWOZ dataset, impacting the performance of models for DST and response generation tasks.  
\end{itemize}

\section{Methodology}\label{sec:methods}

\subsection{MultiWOZ 2.2 Dataset}
The MultiWOZ 2.2 \cite{zang-etal-2020-multiwoz} dataset is an improved version of MultiWOZ 2.1 \cite{eric-etal-2020-multiwoz} correcting DST annotations of 17.3\% of utterances and ontology issues associated with some of the slot values. The dataset contains 10,437 conversations divided between train (8,437), validation (1,000), and test (1,000) sets. Each conversation contains alternating turns between the user and the system utterances. Although the dataset comprises 7 domains, the domains of hospital and police are only present in the training data. Following the work of \cite{lee2021dialogue, wu-etal-2019-transferable}, we remove the conversation of these domains from the training set. The domains of restaurant, hotel, taxi, and train are bookable as the users can ask the system about booking these for them. The dataset also contains an external database for each domain, which contains a list of entries and their attributes. The system responses are supposed to suggest and offer a reservation for only these entries. Ground truth data comprises detailed annotations of dialog states, system responses, and conversation goals, which are used to compare and evaluate the model-predicted responses.

\subsection{Scoring Metrics}
We follow the standard and the widely used metrics for the tasks of DST and Response Generation presented in the work\footnote{\url{https://github.com/Tomiinek/MultiWOZ_Evaluation}} \cite{nekvinda-dusek-2021-shades}. 

\textbf{DST}  The DST performance is measured by Joint State Accuracy (JSA) between the predicted and the ground-truth states. Each slot consists of a triplet of (domain, slot key, and slot value), and multiple such slots can be present in a given dialog history. Some of these slots like \texttt{hotel-stars} are categorical in nature so can take only a fixed set of values while others like \texttt{hotel-name} can take any value based on conversation history. The binary JSA values are computed at each user turn of the conversation and involve comparison between all predicted states and the ground truth states. The domain name and slot key must match exactly, but a fuzzy margin is applied when comparing slot values to ignore minor syntactic differences.

\textbf{End-to-end Response Generation}  The overall TODS pipeline, including DST and response generation, is evaluated using Inform, Success, and BLEU scores. The \textbf{Inform} rate is a conversation-level binary variable that indicates whether the user is presented with the correct venues according to their constraints. For example, for the conversation in Figure \label{fig:overview}, if the system provides the user with a restaurant located in the center of town that serves Chinese food, it will be considered a correctly informed conversation. A multi-domain conversation will be considered correctly informed if the right venues are presented in each domain. Furthermore, a conversation has a \textbf{Success} rate of one if, firstly, the conversation has an inform rate of one (the user was provided the right venue) and secondly, the user was presented with the right attributes about the venue. For instance, in the above example, if the user was provided with the correct restaurant and later during the conversation, the user asks about the restaurant's phone number, postcode, etc., the conversation would be considered successful if this information was presented to the user and unsuccessful otherwise. Like the inform rate, success metric is also binary and is computed at a conversation level. To evaluate the quality of the generated text, \textbf{BLEU scores} are calculated between delexicalized predicted responses and ground truth system responses. This delexicalization process ensures that model predictions are not unfairly penalized when they provide information about a suitable venue that differs from the specific venue chosen in the ground truth response, as multiple venues may satisfy user constraints. The following definition of the combined score is used to compare approaches in the MultiWOZ benchmark.

$$\text{Combined Score} = \frac{(\text{Inform} + \text{Success})}{2} + \text{BLEU Score}$$

Additionally, some metrics to compare textual richness are also compared.

\subsection{Approach}

\subsubsection{DST Prediction}
We start by building a DST pipeline for MultiWOZ, which takes the dialog history as input and outputs the slots present in the context. We run all of our experiments with Flan-T5-large, Mistral-7B, and Claude Sonnet 3.0 as our base models, as they represent a comprehensive range of model sizes, spanning from sub-1B parameters to mid-sized 7B models and larger LLMs, allowing us to assess performance across varying scales. 

\textbf{Single fine-tuned Agent} This approach fine-tunes a language model to generate the list of slots given the input. We used the models Mistral-7B and Flan-T5-large for this approach. For model training, we treat each segment of the dialogue history up to each user utterance as a separate example, instead of updating a running dialogue state with each new utterance. This approach offers two key advantages: i) It prevents the accumulation of errors that could occur with subsequent dialogs ii) It better handles scenarios present in MultiWOZ datasets where previous dialogue slots are cleared, a situation where continuously updating dialogue states would fail.

\textbf{Prompted LLM} In this approach, we presented the Claude Sonnet 3.0 model with detailed instructions for the DST task. These instructions include a list of possible slots to track, all possible values categorical slots can take, explicit output format, and some in-context examples from which the models can learn. We present Claude with 50 random in-context examples selected at random from the training set.

\textbf{Multi-Agent DARD}
In this approach, we fine-tune domain-specific DST models. We train a separate distinct model for each domain, focusing exclusively on tracking the slots relevant to that domain. To prepare the training data, we segregate the slots by domain; for instance, if a training sample contains slots from both the attraction and train domains, we add that sample to the training data of both domain models, with outputs consisting of slots from the respective domain only. During the testing phase, we first pass the context to a dialog manager agent, which outputs all domains whose slots are present. We then invoke the respective domain models to obtain the final set of slots. We use a prompted Claude 3.0 Sonnet LLM as a dialog manager agent to inform us about the domains present in the dialog context.

\subsubsection{Response Generation Pipeline}
The response generation pipeline involves predicting the delexicalized system utterance given the conversation context up to the previous user utterance. Similar to the DST pipeline, we choose to experiment with a single-agent approach and a domain-specific multi-agent approach. 

\textbf{Single fine-tuned Agent} This approach fine-tunes a single model to handle conversations across all domains, providing it with both the dialog context and details of venues meeting user criteria. Appendix \label{resgen_example} presents an example of the input provided to the model. These venue details, obtained by querying the database using predictions from the best DST pipeline, include the number of matching venues and specifics of one of them, if any. Adding these venue details allows the model to tailor responses based on the number of available options, whether suggesting a single match, asking for more preferences when multiple venues fit, or informing the user when no matches are found.

\textbf{Multi-Agent DARD} For this approach, we again use models that are specialized to respond to queries from specific domains. We experimented with fine-tuned models(Mistral-7B \& Flan-T5-Large) and instructed Claude Sonnet 3.0 as our domain agents. The Claude-based agent is prompted with detailed information about the list of possible delexicalized tags to be used, generic instructions on how to respond, and some in-context examples. Appendix \label{restaurant_booking_example} presents the prompt used for restaurant agents. The generic instructions on how to respond were designed using manual observation of training data from each domain. We use the Sonnet 3.0 dialog manager agent to determine which domain agent will be best equipped to respond based on the conversation context. We then pass the conversation context and venue details to the delegated agent to generate the system response. Since we can choose any type of agent for each domain, we also experiment with and select the best-performing agent from among Claude Sonnet 3.0, Flan-T5-Large, and Mistral-7B agents based on their combined scores on the validation set.

\section{Results} \label{sec:results}

Table \label{tab:DST_results} compares the results of our DST experiments with the existing best-performers \cite{su2021multitask, niu2024enhancingdialoguestatetracking, feng2023llmdrivendialoguestatetracking}. We divide the existing work primarily into those that use fine-tuned models and those that use prompting LLM methods. We observe that for the fine-tuned Flan-T5-Large model, the performance is much better with domain-specific agents than with a single agent. However, for a fine-tuned Mistral-7B, the performance is nearly the same with the two approaches, and the single model performs a little better. We notice that Claude's performance was poorer than that of the fine-tuned models. Our approach of using a single fine-tuned Mistral-7B model performs better than most existing works, second only to the method followed by \cite{niu2024enhancingdialoguestatetracking}. \cite{niu2024enhancingdialoguestatetracking} fine-tune a LLaMA-2 7B model on the DST dataset, first on the original training data itself(LUAS$_{R}$), which leads to similar performance as that of fine-tuned Mistral-7B, and then on both original training data and augmentations
(LUAS$_{R+G}$), producing marginally better performance than a fine-tuned Mistral-7B models. The prompt-based approaches described in \cite{feng2023llmdrivendialoguestatetracking} performed better than our prompting-based Claude Sonnet method. We believe this stark difference in performance is due to the fact that the work in \cite{feng2023llmdrivendialoguestatetracking} employs a single-slot return approach, where the model predicts each slot individually for each conversation context, resulting in 30 calls in total for each sample's complete prediction. This approach simplifies the task for the LLM, as it only needs to check for one slot in the context at a time, unlike our method, which requires consideration of all possible slots. However, the single-query method requires roughly 30 times more LLM calls.

Table \ref{tab:res_gen_compare} shows the evaluation results of the end-to-end response generation pipeline of the methods that we tried and also some of the best-performing models on the benchmark. We present the evaluation metrics of Inform, Success, and BLEU scores, along with the overall comparison of combined scores. As can be seen from the table, the inform and success rates of Claude-based agents are much better than those of existing works and fine-tuned agents; however, the opposite is true for the BLEU scores. Since the multi-agent framework allows the composability of using different agents for different domains, we select the best-performing agents based on domain-wise performance on the validation set. This combination achieves better information, success rate, and combined score than the existing works. In terms of textual richness measured by Conditional Bigram Entropy (CBE), \#unique words, and \#unique 3 grams, we observe that Claude-based agents are much more lexically diverse than fine-tuned models and other works. We present a detailed discussion of the results in the next section.

\begin{table}[t]
    \centering
    \small
    \setlength{\tabcolsep}{8pt}
    \begin{tabular}{l c}
    \toprule
    Method & Joint Slot Accuracy \\
    \midrule
    \rowcolor[gray]{.9}
    \multicolumn{2}{l}{\textbf{Existing Works - Fine Tuned}} \\
    SDP-DST \cite{lee2021dialogue} & 57.3 \\
    TOATOD \cite{su2021multitask} & 63.79 \\
    D3ST \cite{zhao2022descriptiondriventaskorienteddialogmodeling} & 57.8 \\
    DAIR \cite{huang2023robustnessdataaugmentationloss} & 59.9 \\
    LUAS$_{R}$ \cite{niu2024enhancingdialoguestatetracking} & 65.4 \\
    LUAS$_{R+G}$ \cite{niu2024enhancingdialoguestatetracking} & \textbf{66.3} \\
    \midrule
    \rowcolor[gray]{.9}
    LLaMa-7B \cite{feng2023llmdrivendialoguestatetracking} & 55.37 \\
    LDST \cite{feng2023llmdrivendialoguestatetracking} & 60.65 \\
    \midrule
    \rowcolor[gray]{.9}
    \multicolumn{2}{l}{\textbf{Fine Tuned Flan-T5-Large }} \\
    Single Agent  & 58.9 \\
    Multi-Agent DARD & 63.6 \\
    \midrule
    \rowcolor[gray]{.9}
    \multicolumn{2}{l}{\textbf{Claude Sonnet 3.0 }} \\
    Instruction Prompted & 45.8 \\
    \midrule
    \rowcolor[gray]{.9}
    \multicolumn{2}{l}{\textbf{Fine Tuned Mistral-7B}} \\
    Single Agent & 66.0 \\
    Multi-Agent DARD & 63.1 \\
    \bottomrule
    
    \end{tabular}
    \caption{In this table, we compare our DST approach with the top-performing existing models on the benchmark. * - Numbers for these models are taken from the work of \cite{feng2023llmdrivendialoguestatetracking}}\label{tab:DST_results}
\end{table}

\begin{table*}[t]
    \centering
    \small
    \setlength{\tabcolsep}{4pt}
    \begin{tabular}{l c c c c c c c c}
    \toprule
    Method & BLEU & Inform & Success & Combined Score & CBE & \# words & \# 3-grams \\
    \midrule
    \rowcolor[gray]{.9}
    \multicolumn{8}{l}{\textbf{Existing Works}} \\
    GALAXY \cite{he2022galaxy} & 19.64 & 85.4 & 75.7 & 100.2 & 1.75 & 295 & 2275 \\
    TOATOD \cite{su2021multitask} & 17.04 & 90.0 & 79.8 & 101.9 & - & - & - \\
    RewardNet \cite{feng2023fantasticrewardstamethem} & 17.6 & 87.6 & 81.5 & 102.2 & 1.99 & 423 & 3942 \\
    Mars \cite{sun2023marsmodelingcontext} & \textbf{19.9} & 88.9 & 78.0 & 103.4 & 1.65 & 288 & 2264 \\
    KRLS \cite{yu2023krlsimprovingendtoendresponse} & 19.0 & 89.2 & 80.3 & 103.8 & 1.90 & 494 & 3884 \\
    DiactTOD \cite{wu2023diacttodlearninggeneralizablelatent} & 17.5 & 89.5 & 84.2 & 104.4 & 2.00 & 418 & 4477 \\
    \midrule
    \rowcolor[gray]{.9}
    \multicolumn{8}{l}{\textbf{Fine Tuned Flan-T5-Large}} \\
    Single Agent & 13.0 & 51.1 & 44.5 & 60.8 & 1.7 & 383 & 2658 \\
    Multi-Agent DARD & 15.6 & 82.8 & 70.7 & 92.3 & 1.82 & 354 & 2871 \\
    \midrule
    \rowcolor[gray]{.9}
    \multicolumn{8}{l}{\textbf{Claude Sonnet 3.0}} \\
    Multi-Agent DARD & 9.5 & 95.6 & 88.0 & 101.3 & 2.37 & \textbf{1197} & \textbf{13742} \\
    \midrule
    \rowcolor[gray]{.9}
    \multicolumn{8}{l}{\textbf{Fine Tuned Mistral-7B}} \\
    Single Agent & 15.6 & 81.0 & 63.6 & 87.9 & 2.75 & 930 & 12552 \\
    Multi-Agent DARD & 15.2 & 78.8 & 61.2 & 85.2 & \textbf{2.79} & 993 & 13317 \\
    \midrule
    \rowcolor[gray]{.9}
    \multicolumn{8}{l}{\textbf{Domain Selective Agents}} \\
    Best Domain Agent - DARD & 12.1 & \textbf{96.6} & \textbf{88.3} & \textbf{104.6} & 2.33 & 1098 & 10991 \\
    \bottomrule
    \end{tabular}
    \caption{This table shows the performance comparison between our approaches vs the best-performing models on the End-to-end response generation pipeline. The last three metrics indicate the textual richness. CBE denotes the conditional bigram entropy, \# words and \# 3-grams are the number of unique words and trigrams, respectively.}
    \label{tab:res_gen_compare}
\end{table*}

\section{Discussions}\label{sec:discussions}

In this section, we compare and present some key insights based on various experiments that we tried.
\subsection{Analysis of Claude's DST performance} We conducted a thorough analysis of errors made by Sonnet 3.0 and discovered that most mistakes stemmed from its tendency to track slots in both user and system utterances, whereas the ground truths typically only include slots from user utterances. Table \label{tab:claude-jsa-errors} categorizes these errors. We define over-prediction as cases where groundtruth slots are a subset of predicted slots, under-prediction where groundtruth slots are a superset of predictions, both mismatch when neither is a subset or superset of the other, and value match error when predictions have an incorrect value for any slot. The table reveals that in over half (54\%) of cases, predicted responses contain more slots than the groundtruth. Examining about 100 random samples from this set showed that this occurs because Sonnet responses track slots from system utterances as well. An example of this is provided in Appendix \label{claude_dst_err}. Further investigation uncovered annotation inconsistencies in the MultiWOZ 2.2 dataset itself, with some annotators only tracking slots from user utterances while others included slots from both user and system utterances. This issue is also highlighted in \cite{qian2022annotationinconsistencyentitybias}, which offers a detailed analysis of the extent of this problem in the MultiWOZ 2.2 dataset. To address this inconsistency, \cite{qian2022annotationinconsistencyentitybias} corrected the DST labels in over 70\% of the dialogs in the dataset 

\begin{table}[h]
\centering
\begin{tabular}{|l|c|}
\hline
\textbf{Error Type} & \textbf{Percentage} \\
\hline
Over-prediction & 54\% \\
Under-prediction & 11\% \\
Both Mismatch & 13\% \\
Value Match Error & 22\% \\
\hline
\end{tabular}
\caption{Classification of the errors made by Claude Sonnet 3.0}
\label{tab:claude-jsa-errors}
\end{table}

\subsection{Single Agent Vs Multi-Agent Approaches}

\textbf{DST} We observe that the JSA of the domain-specific fine-tuned Flan-T5-Large model is 4.6\% better than that of a single model. However, for the fine-tuned Mistral-7B, the difference is not as big, and the single-agent approach performs better. We attribute this observation to the fact that Mistral-7B is a more powerful and larger model ($\sim$ 7B params) compared to Flan-T5-Large ($\sim$780M params), hence it is more capable of modeling slots of all domains in a single model. While having domain specialization is useful for the smaller Flan model, it brings no advantages for the bigger Mistral-7B model. A possible drawback of the domain-specific agents is that they are trained on a smaller amount of data than a single agent, and they also do not support cross-domain transfer learning.

\textbf{Overall TODS} Just like the DST pipeline, we get similar observations for the fine-tuning-based approaches, i.e., we get large (31.5\%) improvements with domain-specific agents for Flan-T5-Large while Mistral-7B sees a slight decrease (2.7\%) in performance. In terms of textual richness, we notice that multi-agent approaches have great lexical diversity. Intuitively, this aligns with our expectations as domain-specific agents are more likely to learn and use domain-specific corpora, leading to higher overall diversity.

Overall, the DARD-based multi-domain approach offers significant flexibility and composability for TODS. This framework serves as a versatile plug-and-play environment for various domain agents, allowing us to select the most effective agents for each domain. For instance, in the case of the MultiWOZ dataset, Claude 3.0 Sonnet worked best for the attraction, hotel, and restaurant domains, while Mistral-7B worked best for the train domain, and Flan-T5-Large worked best for the taxi domain. The multi-agent structure improves interpretability and simplifies the development of improved domain-specific agents. Throughout the development process, we can monitor domain-wise accuracies, focusing on improving underperforming domains without affecting others. Additionally, the domain-expert approach enables runtime performance optimization by employing smaller, faster models for simpler domains while utilizing larger language models for more complex ones. However, these advantages rely on the crucial assumption that the dialog manager agent can accurately assign the appropriate agent to each task. While this assumption held true for the MultiWOZ dataset due to its distinct and exclusive domains, it may not universally apply to TODS with overlapping domains in real-world scenarios.

\subsection{Fine-tuned Vs Prompted Models} The BLEU score of prompted Claude is considerably lower compared to that of fine-tuned models and existing works. However, the opposite is true for the inform and success rates. The low BLEU score of Claude's responses can be explained by the fact that LLM-generated responses follow different speaking styles and vocabulary. Additionally, the model prompt contains only 8-10 (<1\%) in-context examples from the training data, while the fine-tuned models are trained on the entire dataset (100\%). Hence, the fine-tuned models have a better understanding of responding as an annotator resulting in a higher BLEU score. While the responses from prompted LLMs may have low BLEU scores these responses are preferred by human evaluators  \cite{li2023guidinglargelanguagemodels} \cite{xu-etal-2024-rethinking}. These studies suggest that the low BLEU scores is due to lack of grounding with the dataset.
Appendix \label{claude_res_gen_example} presents an example of a conversation using Claude-based agents with a low BLEU score that nonetheless demonstrates reliable inform and success rates.

We also analyzed conversations where fine-tuned agents failed to achieve inform and success rates, but Claude agents succeeded. In 52\% of these cases, the fine-tuned model failed to inform because none of its responses contained any suggested venues. We found that this issue arises due to another annotator disagreement in the dataset: while some annotators preemptively suggest venue names in earlier utterances, others ask users about additional preferences first before suggesting them with a particular venue. Figure \label{fig:annotator_dis} illustrates this discrepancy. 
\begin{figure*}[htp]
    \centering
    \includegraphics[width=\textwidth]{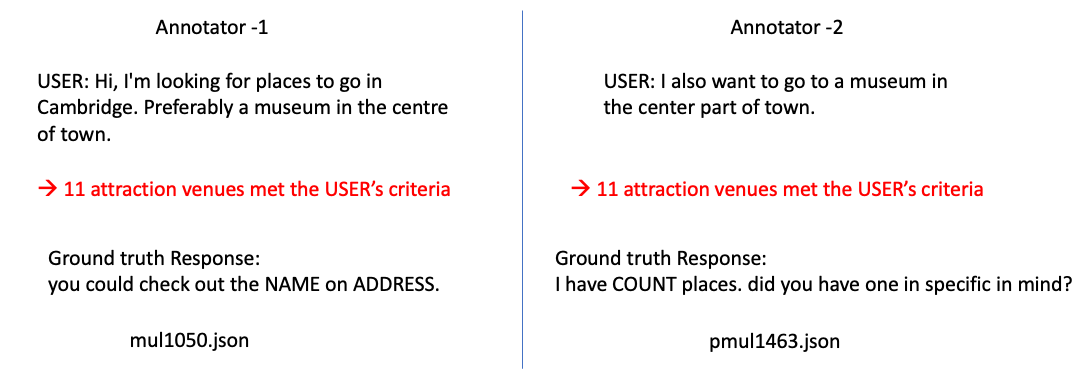}
    \caption{Two samples from the MultiWOZ 2.2 dataset which show how different annotators respond to the users}
    \label{fig:annotator_dis}
\end{figure*}
For the same query (museums in the center) with 11 matching venues, annotator-1 immediately recommended a specific venue, while annotator-2 informed the user about the available choices and asked if they had a particular venue in mind. Both methods are valid, but fine-tuned models sometimes adopt the preference-asking way when the ground truth expects an immediate suggestion. As a result, in subsequent interactions, the model assumes a venue name was already mentioned based on the ground truth and doesn’t repeat it. Consequently, no responses contain a venue name, leading the evaluation system to conclude that the agent failed to suggest any venues. Table \label{tab:venue-suggestions} presents this variation in annotator behavior. For a larger number of venues, most annotators prefer asking for further preferences before making a suggestion, but some directly suggest a name. This often causes the fine-tuned models to miss out on making any suggestions. This issue does not occur with Claude agents, as they are prompted to do both: make a suggestion while also asking for user preferences.

\begin{table}[h]
\centering
\begin{tabular}{|l|c|}
\hline
\textbf{Num. Eligible Venues} & \textbf{Venue Suggested} \\
\hline
Large ($\geq$10) & 24.5\% \\
Medium ($\geq$5 and $<$10) & 35.4\% \\
Small ($<$5) & 55.6\% \\
\hline
\end{tabular}
\caption{Percentage of times when the annotators immediately recommended user a venue, based on different numbers of venues}
\label{tab:venue-suggestions}
\end{table}
\vspace{-15pt}
\section{Conclusion}
In this work, we propose a DARD (Domain Assigned Response Delegation), a multi-agent framework to build Task-Oriented Dialogue Systems (TODS). The DARD framework includes a central dialog manager agent that assigns tasks to domain-specific agents, which then handle and solve these tasks. We evaluate this approach using the Dialogue State Tracking (DST) and Response Generation tasks of the MultiWOZ dataset. For our domain agents, we utilize Flan-T5-Large, Mistral-7B, and Claude Sonnet 3.0 models.

Our findings show that the effectiveness of the multi-agent DARD depends on the type of model used. Smaller models, like Flan-T5-Large, experience significant performance improvements with the multi-agent framework, while larger models, such as Mistral-7B, see a slight decrease in performance. Despite this, DARD offers notable benefits in terms of flexibility, composability, and interpretability, making it a valuable framework for developing more efficient TODS. We also observe that while Claude-based agents achieve state-of-the-art inform and success rates, they have lower BLEU scores due to a lack of alignment with the training samples. However, Claude-generated responses are much more lexically diverse than those from other models. Additionally, our work highlights two significant annotator discrepancies in the MultiWOZ 2.2 dataset. The first discrepancy involves the tracking of dialog slots: while some annotators track only the slots from user utterances, others track slots from both user and system utterances. This inconsistency leads to overprediction when using language models like Claude for Dialogue State Tracking (DST). The second discrepancy relates to the conversation policy. Some annotators preemptively suggest the name of a possible venue to the user, while others first ask for user preferences before suggesting a name. This inconsistency causes fine-tuned models to assume that a venue name was already suggested in a previous utterance, leading them not to mention any name in their predicted responses. As a result, these models often fail to achieve high inform and success rates in conversations. 

\section{Future Work \& Limitations}

While our work tests DARD on the MultiWOZ benchmark, additional evaluations on more complex task-oriented dialogue (TOD) datasets, such as the Schema Guided Dataset \cite{rastogi2020scalablemultidomainconversationalagents} and SMD \cite{eric2017keyvalueretrievalnetworkstaskoriented}, are needed for more conclusive insights into the advantages of multi-agent frameworks.
Moreover, it would be beneficial to evaluate DARD-based TOD systems within an interactive evaluation framework. Currently, model responses are assessed statically based on the existing conversation, which can be problematic since user messages often correspond to prior ground truth system messages rather than the predicted responses.
To enhance DARD's performance, one possible approach is to provide domain agents with selective context instead of the entire conversation. In multi-domain scenarios, such as switching from the attraction domain to the taxi domain, the inclusion of irrelevant prior utterances may hinder the new agent's performance. A dialog manager could mitigate this issue by determining the active domain and selecting only the relevant context.



\newpage
\bibliographystyle{unsrt}
\bibliography{neurips_2024}






\appendix

\section{Appendix / supplemental material}

\subsection{Example Input provided for Response Generation} \label{resgen_example}

\fcolorbox{black}{white}{
\begin{minipage}{0.95\columnwidth}
\textbf{USER}: I need a place to dine in the center thats expensive\\
\textbf{SYSTEM}: I have several options for you; do you prefer African, Asian, or British food?\\
\textbf{USER}: Any sort of food would be fine, as long as it is a bit expensive. Could I get the phone number for your recommendation?\\

Number of restaurants that meet the user's criteria: 33\\
One of them is the following:\\
  <restaurant>\\
   address - 106 Regent Street City Centre\\
   area - centre\\
   food - indian\\
   id - 19214\\
   name - curry garden\\
   phone - 01223302330\\
   postcode - cb21dp\\
   pricerange - expensive\\
   type - restaurant\\
  </restaurant>\\
\end{minipage}
}


\subsection{Claude DST Response Error}\label{claude_dst_err}
\fcolorbox{black}{white}{
\begin{minipage}{0.95\columnwidth}
Conversation Context:\\
\textbf{USER}: My friends said that I must visit Corpus Christi, can you please give me some more information about it?\\
\textbf{SYSTEM}: I certainly can. Corpus Christi is a college located in the \textcolor{red}{center} of town, on King's Parade. There's a two pound entrance fee. Do you need any additional information?\\
\textbf{USER}: Yes. What is the postcode?\\
\textbf{SYSTEM}: The postcode is cb21rh. Do you need anything else?\\
\textbf{USER}: Are there any African restaurants in the centre?\\
\textbf{SYSTEM}: I have one listing for bedouin would that work for you?\\
\textbf{USER}: Is this listing in the expensive price range?\\
\\
Ground-Truth Slots:\\
\textbf{restaurant}: {area: centre, food: african, name: bedouin, pricerange: expensive}\\
\textbf{attraction}: {name: corpus christi, type: college}\\

Claude Predicted Slots:\\
\textbf{restaurant}: {area: centre, food: african, name: bedouin, pricerange: expensive}\\ 
\textbf{attraction}: {\textcolor{red}{area: centre}, name: corpus christi, type: college}
\end{minipage}
}

\subsection{Claude Response Generation Example Conversation}\label{claude_res_gen_example}
\fcolorbox{black}{white}{
\begin{minipage}{0.95\columnwidth}
\textbf{USER}: I would like to try saigon city.\\
\textbf{Ground truth}: great choice. when can i make the reservation for?\\
\textbf{Claude Response}: NAME is an FOOD restaurant located at ADDRESS in the AREA area. their phone number is PHONE. would you like me to book a table for you?\\
\textbf{BLEU}: 1.76\\

\textbf{USER}: 5 people on a Wednesday at 19:45.\\
\textbf{Ground truth}: you got it. here is your reference number: REFERENCE.\\
\textbf{Claude Response}: booking confirmed for NAME on DAY at TIME for COUNT people. your reference number is REFERENCE.\\
\textbf{BLEU}: 10.93\\

\textbf{USER}: Thank you for help. That's everything I needed.\\
\textbf{Ground truth}: Have a great day!\\
\textbf{Claude Response}: you're welcome. thank you for contacting us!\\
\textbf{BLEU}: 4.76\\
\textbf{Avg. Conversation BLEU}: \textcolor{red}{5.81}
\end{minipage}
}

\subsection{Fine-Tuned Models didn't offer any suggestion}
\fcolorbox{black}{white}{
\begin{minipage}{0.95\columnwidth}
\textbf{USER}: I am coming to visit and will be in the centre area. I would like an attraction to visit.\\
\textbf{SYSTEM}: sure, how about NAME?\\
\textbf{Predicted Response}: there are COUNT attractions in the AREA. what type of attraction are you interested in?\\
\\
\textbf{USER}: Sounds interesting! What is the address?\\
\textbf{SYSTEM}: Located at ADDRESS .\\
\textbf{Predicted Response}: the address is ADDRESS\\
\\

\textcolor{red}{None of the predicted response present any museum}
\end{minipage}
}

\subsection{Claude Based Restaurant Agent Prompt}\label{restaurant_booking_example}
\fcolorbox{black}{white}{
\begin{minipage}{0.95\columnwidth}
\textbf{<instructions>}\\
You are supposed to act a system that assists users with their queries about finding and booking restaurants.\\
You will be provided with the following information:\\
- Conversation History between the USER and the SYSTEM\\
- Number of restaurants that meet the user's requirements. You will also be provided with details of one of the restaurant that meet the user's requirements, if any\\
You have to generate the following:\\
    - Delixicalized response to the last USER message\\
    - Values of the delexicalized tokens you used in your response\\
\textbf{</instructions>}\\
\\
\textbf{<Delexicalization>}\\
You are allowed to use the following delexicalized tokens:\\
1. [restaurant\_name] - for the name of the restaurant\\
2. [restaurant\_food] - for the food cuisine of the restaurant\\
3. [restaurant\_pricerange] - for the price range of the restaurant\\
4. [restaurant\_area] - for the area of the restaurant\\
5. [restaurant\_address] - for the address of the restaurant\\
6. [restaurant\_phone] - for the phone number of the restaurant\\
7. [restaurant\_postcode] - for the postcode of the restaurant\\
8. [restaurant\_choice] - used at instances where you provide user with a choice of multiple restaurants\\
9. [restaurant\_ref] - used to provide the booking reference to the user\\
10. [restaurant\_booktime] - used to provide the booking time to the user\\
11. [restaurant\_bookday] - used to provide the booking day to the user\\
12. [restaurant\_bookpeople] - used to provide the number of people you have booked the restaurant for\\
\textbf{</Delexicalization>}\\
\\
\textbf{<how to respond>}\\
These are the specific guidelines to follow while responding to the user:\\
1. Do not write very long responses for the user it should be at max 1 to 2 lines\\
2. Always inform the user about the things they asked for in their last utterance, it can be things like phone number, address, postcode, etc\\
3. Whenever you book a restaurant you should always provide the booking reference to the user, also inform that their table will be held for 15 minutes\\
4. In the instances where multiple restaurants meet user's requirements, you should inform the user about the count using [restaurant\_choice] token and also try to suggest user one of the restaurant using [restaurant\_name] token\\
5. Before you book the restaurant be sure to ensure that the user has provided the information about bookpeople, bookday and booktime. If they haven't first ask them to provide these information\\
6. If you notice that none of the restaurant meet user's criteria, just politely inform them that no restaurant is found that satisfies the user's query\\
7. If the user's done with their query, just say thanks for contacting us and end the conversation.\\
8. You may encounter cases where the restaurant that has been suggested to the user in the conversation has not been provided to you as example this will be because you are only presented with one of the examples. In these cases you have to respond normally to the user as if you know infomration about the suggested reataurant \\
\textbf{</how to respond>}\\
\\

\end{minipage}
}
\newpage
\fcolorbox{black}{white}{
\begin{minipage}{0.95\columnwidth}
\textbf{<output format>}\\
You have to strictly follow the following format while generating the response:\\
Response: The delixicalized response to the last USER utterance\\
Token\_values: [delixicalized tokens] - [value], ....\\
Reasoning: The reasoning behind the response\\
\textbf{</output format>}\\

\textbf{<examples>}\\
In this section you will be presented with some of the examples you can learn from:\\
\\
\textbf{<examples\_about\_suggesting\_restaurants>}\\
This subsection will provide you examples of how to suggest restaurants to the user.\\
.....\\
\textbf{</examples\_about\_suggesting\_restaurants>}\\
\\
\textbf{<examples\_no\_restaurant\_found>}\\
This subsection contains examples where no restaurant was found meeting user's criteria\\
...\\
\textbf{</examples\_no\_restaurant\_found>}\\
\\
\textbf{<examples\_about\_booking>}\\
This subsection will contain examples related to user queries about booking restaurants\\
...\\
\textbf{</examples\_about\_booking>}\\
\\
\textbf{<examples\_ending\_conversation>}\\
This subsection will contain examples related to ending the conversation\\
...\\
\textbf{</examples\_ending\_conversation>}\\
\\
\textbf{</examples>}\\

Now its your turn to answer, generate the response to the following conversation history:\\

\end{minipage}

}

\end{document}